\pdfoutput=1

\documentclass[11pt]{article}

\usepackage[preprint]{acl}

\usepackage{times}
\usepackage{latexsym}

\usepackage[T1]{fontenc}

\usepackage[utf8]{inputenc}

\usepackage{microtype}

\usepackage{inconsolata}

\usepackage{graphicx}

\usepackage{amsmath}
\usepackage[capitalize]{cleveref}
\Crefname{section}{Sec.}{Secs.}
\Crefname{equation}{Eq.}{Eqs.}
\Crefname{appendix}{App.}{Apps.}
\Crefname{figure}{Fig.}{Figs.}
\Crefname{table}{Tab.}{Tabs.}
\Crefname{algorithm}{Alg.}{Algs.}
\usepackage{booktabs}
\usepackage{moresize}
\usepackage{color}
\newcommand{\ourmethod}{\texttt{MTPF-TPL}}
\usepackage{tcolorbox}
\tcbuselibrary{breakable}
\usepackage{graphicx}
\usepackage{subcaption}
\usepackage{soul,xcolor}               
\sethlcolor{yellow!40}                 
\newcommand{\ent}[1]{\hl{#1}}          
\usepackage{siunitx}
\usepackage{multirow}
\usepackage{hhline}
\usepackage{pifont}
\newcommand{\bftab}{\fontseries{b}\selectfont}
\usepackage{hyperref}

\title{Multi-Task Pre-Finetuning of Lightweight Transformer Encoders for Text Classification and NER}

\author{
Junyi Zhu$^{1}$\thanks{Correspondence to: \{junyi.zhu, savas.ozkan, m.ozay\}@samsung.com}  \quad Savas Ozkan$^{1}$\footnotemark[1]  \quad Andrea Maracani$^{1}$ \quad Sinan Mutlu$^{1}$ \quad \\ 
\textbf{Cho Jung Min}$^{2}$ \quad \textbf{Mete Ozay}$^{1}$\footnotemark[1]
\\
$^1$Samsung R\&D Institute UK (SRUK) \quad
$^2$Samsung Electronics Korea
}

\begin{document}
\maketitle
\begin{abstract}
Deploying natural language processing (NLP) models on mobile platforms requires models that can adapt across diverse applications while remaining efficient in memory and computation. We investigate pre-finetuning strategies to enhance the adaptability of lightweight BERT-like encoders for two fundamental NLP task families: named entity recognition (NER) and text classification. While pre-finetuning improves downstream performance for each task family individually, we find that naïve multi-task pre-finetuning introduces conflicting optimization signals that degrade overall performance. To address this, we propose a simple yet effective multi-task pre-finetuning framework based on task-primary LoRA modules, which enables a single shared encoder backbone with modular adapters. Our approach achieves performance comparable to individual pre-finetuning while meeting practical deployment constraint. Experiments on 21 downstream tasks show average improvements of $+0.8\%$ for NER and $+8.8\%$ for text classification, demonstrating the effectiveness of our method for versatile mobile NLP applications.
\end{abstract}

\section{Introduction}
\label{sec:intro}
Mobile applications such as automatic calendar event creation from emails and personalized recommendations based on messages rely on solving multiple natural language processing tasks, particularly text classification and named entity recognition (NER). Solutions often employ either generative large language models~\citep{wang-etal-2024-improving-text,Constantin2024NuExtract} or BERT-like encoder models~\citep{devlin2019bert}, the latter are better suited for on-device deployment due to the demands of memory and computational efficiency.

\begin{figure}[t]
\centering
\includegraphics[width=0.94\columnwidth]{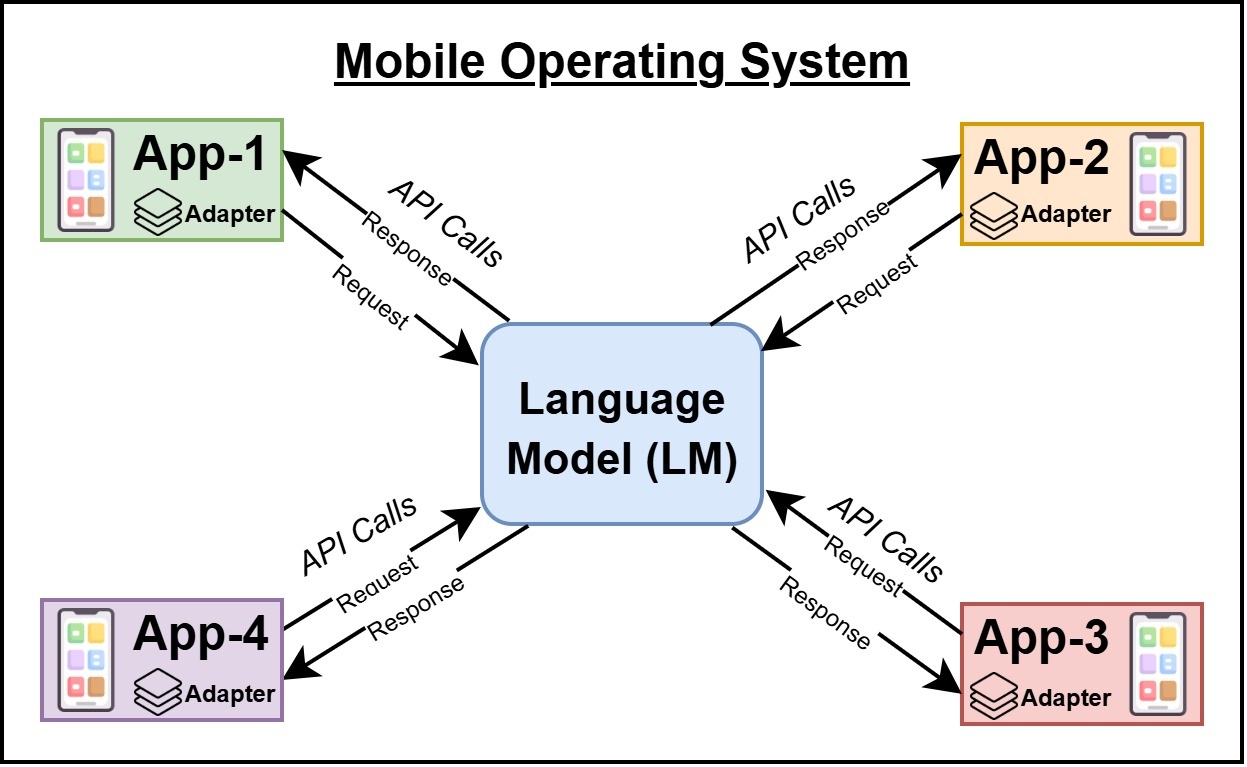}
\caption{\textbf{An illustration of the practical deployment setting on mobile device}. Mobile applications (APP-$\{1\ldots4\}$) use the language model by calling system API and sending their job with task-specific model adapters. Adapters are often in form of LoRA or linear classifier.}
\label{fig:setting}
\end{figure}

As illustrated in \cref{fig:setting}, mobile operating systems (e.g., Android AICore) deploy a shared language model backbone that can be invoked by applications along with their task-specific adapters, such as Low-Rank Adaptation (LoRA)~\cite{hu2022lora} or linear classifiers. This setup requires a highly generalizable backbone—achieved through adapter-based tuning—since mobile applications are diverse and new ones continue to emerge after system deployment.

Directly fine-tuning a pre-trained BERT-like encoder often yields sub-optimal results, particularly when the available data for an application’s sub-task is limited. This is because the pre-trained representations—optimized primarily through masked language modeling~\citep{devlin2019bert,liu2019roberta}—may not align well with the objectives of downstream tasks. \textit{To address this misalignment, a \textbf{pre-finetuning} stage can be introduced:}
\begin{tcolorbox}[breakable,colback=gray!5, colframe=gray!80, boxrule=0.5pt, arc=2pt, left=5pt, right=5pt, top=5pt, bottom=5pt]
Unlike pre-training, pre-finetuning focuses on objectives that are better \textit{aligned with the target task}, using data that typically includes \textit{task-relevant annotations}. Unlike downstream adaptation, however, pre-finetuning is performed on \emph{general}, \emph{large-scale} datasets, and its objectives are \emph{not limited to specific sub-tasks} such as predicting a fixed label set. This process enables the model to acquire more task-relevant representations, thereby improving its adaptability to downstream tasks.
\end{tcolorbox}

In this work, we investigate pre-finetuning strategies to enhance the adaptability of a pre-trained BERT-like encoder across two downstream task families: NER and text classification. We use the term task families because each comprises multiple sub-tasks spanning diverse domains, label sets, or entity types. As these task families depend on distinct representational characteristics—context-level for text classification and token-level for NER—we first implement and evaluate dedicated pre-finetuning strategies for each \textbf{in \cref{sec:pre-finetuneing}}.

Applying two individual pre-finetuning strategies produces separate encoder models, whereas our deployment setting requires a single shared backbone (\cref{fig:setting}). To address this, we explore multi-task pre-finetuning. However, we find that optimizing the model for one task family can degrade its adaptability to the other. As a result, naïve multi-task pre-finetuning—i.e., multi-task pre-finetuning without proper separation—often fails to match the performance of individual pre-finetuned models. \textbf{In \cref{sec:insight}}, we analyze this issue and identify an inherent incompatibility between the pre-finetuning objectives of NER and text classification.

To address this challenge, we propose a simple yet effective framework based on LoRA modules, introduced \textbf{in \cref{sec:mtl}}, with each module tailored to a specific task family (\cref{fig:arch}). During multi-task pre-finetuning, we apply LoRA modules to the last few transformer layers of a pre-trained encoder, updating each module exclusively with its corresponding task objective. Unlike conventional LoRA, which freezes the backbone, we allow the entire encoder to be jointly updated by both objectives. After pre-finetuning, the encoder is deployed within the mobile operating system, while the LoRA modules are distributed to applications. These modules can be used directly for inference or as initialization for downstream task adaptation. We refer to them as task-primary LoRAs.



\paragraph{Our contributions are:} 
\vspace{-.2em}
\begin{itemize}
    \vspace{-.5em}
    \item We implement two individual pre-finetuning strategies for NER and text classification and demonstrate their effectiveness on improving downstream performance (on average \textcolor{green}{+0.8} across 5 NER sub-tasks and \textcolor{green}{+8.8} across 16 text classification sub-tasks). Notably, we propose a novel pre-finetuning strategy for lightweight encoder models on NER via knowledge distillation (see \cref{sec:pre-finetuneing}).
    
    \item Despite their individual effectiveness, we show that the two pre-finetuning strategies interfere with each other. Through analysis, we provide experimental evidence that reveals contradictory evolutions of representational characteristics, highlighting their incompatible optimization directions (see \cref{sec:insight}).

    \item Building on this analysis, we propose a simple yet effective multi-task pre-finetuning framework using task-primary LoRAs to resolve the conflict between strategies. This approach delivers a single shared backbone and distributed task-primary LoRAs, aligning with our deployment constraints (\cref{fig:setting}), while achieving comparable performance to individual pre-finetuned models (see \cref{sec:mtl}).
\end{itemize}
\section{Related Work}

\paragraph{Text Classification.}
Advances in text classification have been driven by transformer-based models~\citep{vaswani2017attention}, with BERT-like encoders~\citep{devlin2019bert, liu2019roberta} setting benchmarks on tasks such as sentiment analysis and topic classification. Lightweight models like MiniLM~\citep{NEURIPS2020_3f5ee243} and DistilBERT~\citep{sanh2019distilbert} have attracted attention for their efficiency in resource-constrained environments, achieving competitive performance with lower computational cost. Additionally, recent work increasingly focuses on pre-finetuning strategies, particularly weakly-supervised contrastive learning with text pairs, which has proven effective for improving downstream performance~\citep{wang2022text, sturua2024jina, gunther2023jina, zhang2023contrastive}.
\vspace{-.5em}
\paragraph{Named Entity Recognition (NER).}
NER involves identifying entities (e.g., person names, organizations) in unstructured text~\cite{jehangir2023survey}. Unlike text classification, which assigns a single label to the entire input, NER requires token-level predictions. Existing methods fall into two categories: token classification~\cite{devlin2019bert}, which labels each token based on predefined entity types, and span classification~\cite{ye2021packed, zhong2020frustratingly, Aarsen_SpanMarker, zhu2022deep}, which detects and classifies text spans. While span classification often yields better performance, especially for complex or nested entities, it is less efficient due to span enumeration and is thus impractical for resource-constrained scenarios such as mobile devices. Similarly, generative approaches based on LLMs~\cite{wang-etal-2025-gpt,ashok2023promptner} are unsuitable for deployment on mobile devices. Given the cost of entity annotation, recent work has also explored pre-finetuning strategies to boost downstream performance~\cite{liu2021ner,bogdanov-etal-2024-nuner}.

\paragraph{Multi-Task Learning of Text Classification and NER.}
Recent studies have explored multi-task training of models using both document-level and token-level annotations to improve overall performance, primarily in specific domains such as quality estimation~\cite{deoghare2023multi}, news recommendation~\cite{bi2022mtrec}, and sentiment analysis~\cite{fan2022sentiment}. However, a systematic analysis of multi-task pre-finetuning for \textit{diverse} text classification and NER tasks remains largely unexplored. Additionally, \citep{feng-etal-2024-mixture} trains multiple domain-specific LoRA modules and uses a trained routing network to select between them. In contrast, we focus on training a generalizable backbone and reducing the number of parameters proportional to the quantity of downstream tasks.
\section{Evaluation of Downstream Performance}
\label{sec:adapt}
We evaluate pre-finetuning strategies based on their downstream performance, which involves adapting the pre-finetuned model to a diverse set of sub-tasks. In this section, we first describe the evaluation setup. In the following sections, we report average metrics across all sub-tasks, with detailed per-dataset results deferred to the appendices. 

\begin{figure}[t!]
\centering
\includegraphics[width=\columnwidth]{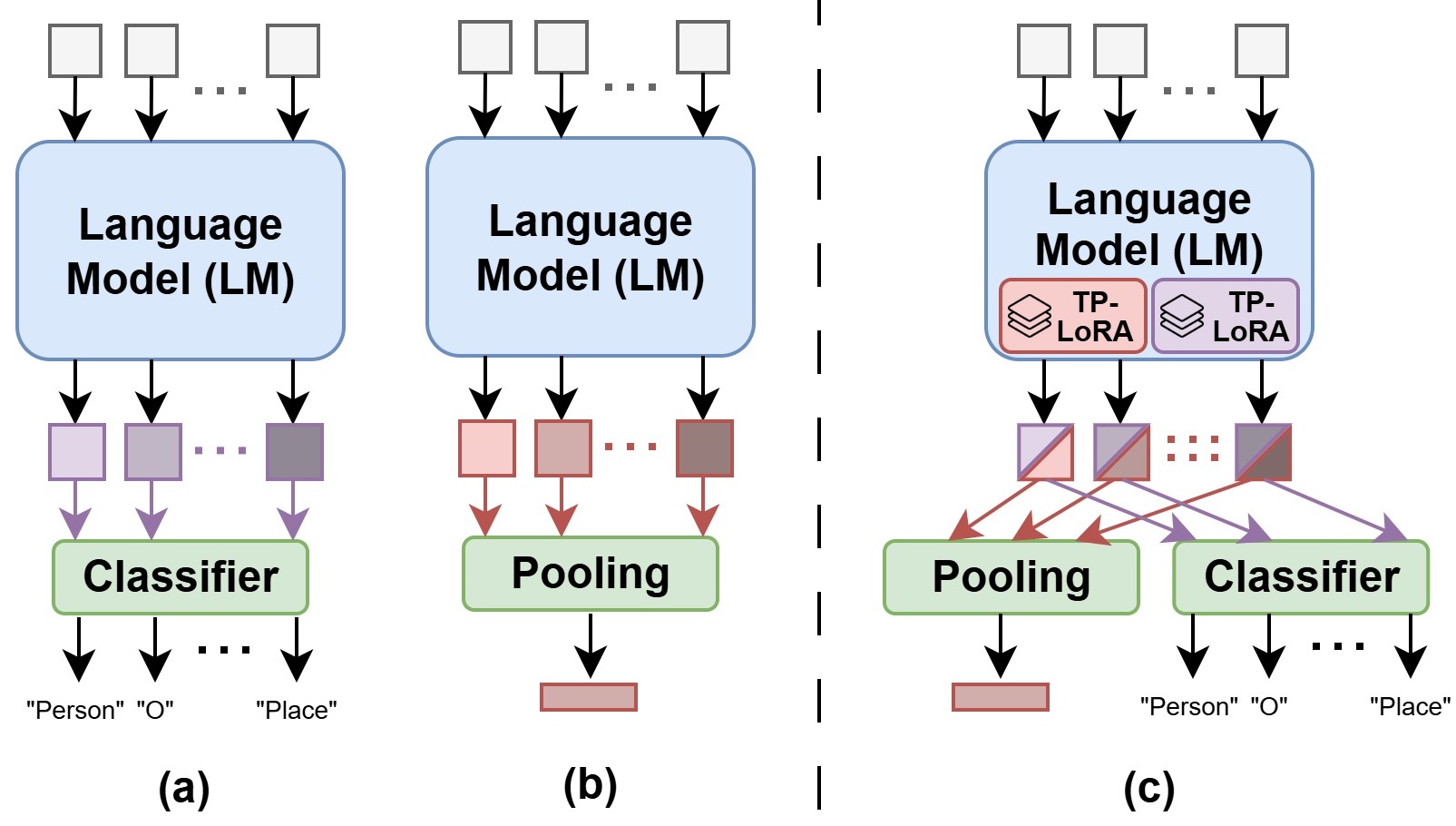}
\caption{\textbf{Illustration of different model configurations for processing inputs}: (a) NER model predicts the entity type of each token; (b) text classification model extracts a text embedding representing the entire input; (c) our proposed shared encoder with task-primary (TP) LoRAs supports diverse outputs, with each LoRA module dedicated to its specific task.}
\label{fig:arch}
\end{figure}

\paragraph{NER Sub-Tasks.} We adopt five NER datasets spanning diverse domains and entity types: \texttt{CoNLL} (4 entity types)\citep{tjong-kim-sang-de-meulder-2003-introduction}, \texttt{MIT Restaurant} (8 types)\citep{6639301}, \texttt{BioNLP} (5 types)\citep{collier-kim-2004-introduction}, \texttt{OntoNotes5} (18 types)\citep{hovy-etal-2006-ontonotes}, and \texttt{CrossNER} (39 types)~\citep{liu2020crossner}.

\paragraph{Text Classification Sub-Tasks.}
We adopt the widely used \texttt{MTEB} benchmark framework~\citep{muennighoff-etal-2023-mteb} and evaluate on 16 datasets covering a variety of domains: \texttt{Banking77}, \texttt{Emotion}, \texttt{AmazonCounterfactual}, \texttt{MassiveIntent}, \texttt{TweetSentiment}, \texttt{ToxicChat}, \texttt{News}, \texttt{Patent}, \texttt{FinancialPhrasebank}, \texttt{FrenkEn}, \texttt{IMDB}, \texttt{ArXiv}, \texttt{DBpedia}, \texttt{TweetTopicSingle}, \texttt{YelpReviewFull}, and \texttt{ToxicConversation}.

\paragraph{Adaptation Setup.}
For text classification sub-tasks, we follow the standard \texttt{MTEB} protocol, training a linear classifier on top of pooled token embeddings extracted from the encoder. We use mean pooling, following the configurations adopted in Jina~\citep{gunther-etal-2023-jina, sturua2024jina, gunther2023jina} and the E5 model family~\citep{wang2022text, wang2024multilingual}.
For NER sub-tasks, we apply LoRA~\citep{hu2022lora} and perform token classification~\citep{devlin2019bert}. Further details of training configurations are provided in \cref{app:adapt}.

\paragraph{Models.}
Motivated by deployment requirements on mobile devices and our stakeholder's interests, we conduct experiments using two lightweight transformer encoders: MiniLM with 33M parameters~\citep{NEURIPS2020_3f5ee243} and DistilBERT with 67M parameters~\citep{sanh2019distilbert}. For both models we consider the English-only version.
\section{Pre-Finetuning Strategies for NER and Text Classification}
\label{sec:pre-finetuneing}
Pre-finetuning is applied to pre-trained models prior to their adaptation to downstream tasks. During pre-finetuning the model is further trained on \textit{general, large-scale datasets with learning objectives that are more closely aligned with the target tasks}. Recent works have demonstrated the effectiveness of pre-finetuning for either NER or text classification in isolation. Accordingly, we adopt and extend these strategies to improve the downstream performance of lightweight encoder models on two task families.

In the remainder of this section, we first describe our pre-finetuning procedures for the two task families, and then analyze the downstream performance achieved by each approach.

\paragraph{Pre-Finetuning for NER.}
\citet{bogdanov-etal-2024-nuner} recently proposed using ChatGPT-3.5 to extract identifiable concepts from large-scale unlabeled corpora such as C4~\citep{JMLR:v21:20-074}, resulting in a dataset of \emph{24.4M words}, \emph{4.38M annotations}, and \emph{200k concepts}. A pre-trained RoBERTa model~\citep{liu2019roberta} was then optimized on this dataset via contrastive learning by aligning entity embeddings with their corresponding concepts, producing the pre-finetuned \texttt{NuNER} model.

However, \texttt{NuNER}’s embedding alignment effectively performs self-distillation, which poses challenges for lightweight models used in on-device applications due to their limited representational capacity. To address this, we distill knowledge from \texttt{NuNER} into lightweight models. Specifically, we adopt \texttt{NuNER}’s 24.4M-word dataset and tag each token using \texttt{NuNER} outputs. Since \texttt{NuNER} lacks a classifier head, we extract token embeddings and apply mini-batch $k$-means clustering to group them, assigning pseudo labels based on cluster IDs. Finally, we train our encoder models—augmented with a linear classifier—to predict these pseudo labels using cross-entropy loss, see \cref{app:pf} for more details.

\begin{table}[t]
\centering
\resizebox{\columnwidth}{!}{
\begin{tabular}{l S[table-format=2.1, table-align-text-post=false]
                   S[table-format=2.1, table-align-text-post=false]
                   S[table-format=2.1, table-align-text-post=false]}
\toprule[1.5pt]
\textbf{Model} & {\textbf{NER (5 DS)}} & {\textbf{TC (16 DS)}} & {\textbf{Average}} \\
\midrule[1.pt]
Base model & 76.4 & 55.1 & 65.8\\
PF for NER & 77.1 \small\textcolor{green}{+0.7} & 53.9 \small\textcolor{red}{-1.2} & 65.5 \small\textcolor{red}{-0.3} \\
PF for TC & 74.6 \small\textcolor{red}{-1.8} & 63.9 \small\textcolor{green}{+8.8} & 69.2 \small\textcolor{green}{+3.4}\\
\bottomrule[1.5pt]
\end{tabular}
}
\caption{\textbf{Comparison of downstream performance across different pre-finetuning (PF) strategies and the base model.} NER results report the average F1 score over 5 datasets (DS), while text classification (TC) results report the average accuracy over 16 datasets. \textcolor{green}{Gain} and \textcolor{red}{Loss} are measured relative to the base model. Results for individual datasets are provided in \cref{tab:pft_ner,tab:pft_tc}. The model architecture used is MiniLM.}
\label{tab:pft}
\end{table}

\paragraph{Pre-Finetuning for Text Classification.}
Following recent work~\citep{wang2022text,gunther2023jina}, we apply \emph{weakly supervised contrastive learning} to pre-finetune the encoder.
Given a paired corpus
$\mathcal{D} = \{(x_i,\, x_i^{+})\}_{i=1}^{N}$,
where each pair shares similar semantics, let $f_\theta$ denote the encoder, and define the $\ell_2$-normalized sentence embeddings as:
\[
\vspace{-0.5em}
z_i = \frac{f_\theta(x_i)}{\lVert f_\theta(x_i) \rVert_2},
\qquad
z_i^{+} = \frac{f_\theta(x_i^{+})}{\lVert f_\theta(x_i^{+}) \rVert_2}.
\vspace{-0.1em}
\]
Contrastive learning encourages the encoder to bring semantically similar pairs closer while pushing apart unrelated (negative) examples.
We adopt \emph{in-batch negatives}: for each anchor $x_i$, its paired text $x_i^{+}$ serves as the sole positive, while all other samples in the minibatch serve as negatives.
The resulting InfoNCE loss~\citep{pmlr-v119-chen20j} is:
\begin{equation}
\vspace{-0.5em}
\label{eq:info_nce}
\nonumber
\mathcal{L}_{\text{CL}} =
-\frac{1}{|B|} \sum_{i \in B}
\log
\frac{\exp\!\big(\langle z_i,\, z_i^{+} \rangle / \tau\big)}
     {\sum_{j \in B} \exp\!\big(\langle z_i,\, z_j^{+} \rangle / \tau\big)},
 \vspace{-0.1em}
\end{equation}
where $\langle \cdot, \cdot \rangle$ denotes the dot product, and $\tau$ is a temperature parameter (set to 0.05).

To ensure data quality and diversity, we compile multiple datasets pre-processed by \texttt{SentenceTransformers}~\citep{reimers-2019-sentence-bert}, totaling \textit{$\sim 895M$} text pairs.
Further dataset and training details are provided in \cref{app:pf}.

\paragraph{Downstream Performance after Pre-Finetuning.}
We evaluate downstream performance after applying each pre-finetuning strategy and compare it to the pre-trained base model. As shown in \cref{tab:pft}, pre-finetuning improves performance within its respective task family, with average gains of \textcolor{green}{$+0.7\%$} for NER and \textcolor{green}{$+8.8\%$} for text classification. Additionally, we observe significant improvement for NER under data scarcity: when only $10\%$ of the original downstream train data is used for adaptation, the pre-finetuned model achieves an average F1 improvement of \textcolor{green}{$+8.4\%$} (see \cref{app:limited_data} for details).

Notably, we observe that each pre-finetuning strategy hinders adaptation to the opposite task family, indicating interference between the two approaches. This effect is consistent across \texttt{NuNER} and other open-source pre-finetuned models for text classification (see \cref{app:open_source} for details). We analyze this issue in the next section and introduce our solution in \cref{sec:mtl}, which reconciles both strategies to pre-finetune a unified encoder.
\begin{figure}[t]
    \centering
    \begin{subfigure}[b]{0.49\linewidth}
        \centering
        \includegraphics[width=\linewidth]{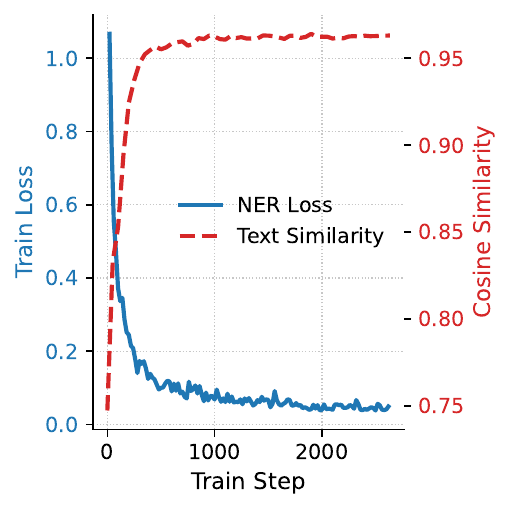}
        \vspace{-2em}
        \caption{Text embedding similarity during NER training.}
        \label{fig:text_similarity}
    \end{subfigure}
    \hfill
    \begin{subfigure}[b]{0.49\linewidth}
        \centering
        \includegraphics[width=\linewidth]{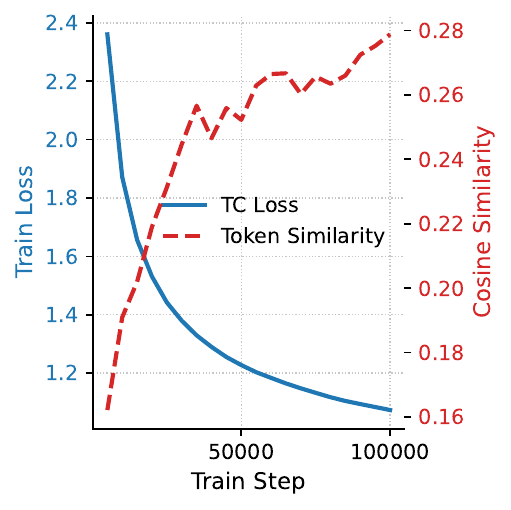}
        \vspace{-2em}
        \caption{Token embedding similarity during TC training.}
        \label{fig:token_similarity}
    \end{subfigure}
    \caption{\textbf{Similarity trends during pre-finetuning}: \textbf{(a)} perturbed sentences, as illustrated in \cref{fig:entity-perturbation}, become closer in embedding space when the model is optimized for NER; \textbf{(b)} token embeddings within the same sentence become more homogeneous when the model is optimized for text classification (TC).}
    \label{fig:insight}
\end{figure}

\section{Interference Between Pre-Finetuning for Text Classification and NER}
\label{sec:insight}
We identify that the interference between pre-finetuning for text classification and NER stems from conflicting requirements on token embeddings. Our analysis reveals two key findings:

\paragraph{Finding 1: Different Sentences Appear Similar After Pre-Finetuning for NER.}
In the NER pre-finetuning setup, a linear classifier operates on token embeddings extracted from the encoder, requiring that tokens representing entities of the same type be mapped to similar representations. As a result, pre-finetuning for NER encourages the model to reduce distinctions between individual entities of the same type.

To examine this effect, we construct a perturbed version of the CoNLL test set. We first extract all entities from the test set and then replace each entity in a sentence with a randomly sampled entity of the same type, generating four perturbed variants per sentence (see \cref{fig:entity-perturbation} for an example).

Next, we train a model on the CoNLL training set for NER and use the resulting encoder to extract text embeddings for both the original and perturbed sentences. We compute the cosine similarity between each perturbed sentence and its corresponding original. As shown in \cref{fig:text_similarity}, similarity increases as training progresses, indicating that the model maps perturbed sentences to increasingly similar representations. This reduction in representational distinctiveness can be detrimental to text classification sub-tasks, which rely on preserving sentence-level differences.

\paragraph{Finding 2: Token Embeddings of the Same Input Become More Homogeneous After Pre-Finetuning for Text Classification.}

In text classification, fine-grained token-level distinctions are less important, as the model focuses on capturing the overall semantics of the input. We find that pre-finetuning for text classification encourages the model to produce more homogeneous token embeddings within each input sequence.

To quantify this effect, we compute pairwise cosine similarities among token embeddings within the same sentence, averaging the results over 2,000 samples throughout pre-finetuning. As shown in \cref{fig:token_similarity}, intra-sentence token similarity increases as training progresses. This trend explains why pre-finetuning for text classification may impair adaptation to NER, which relies on preserving token-level distinctions to classify entities.

\begin{table*}[t!]
\centering
\begin{minipage}[t!]{0.69\textwidth}
\vspace{-.8em}
\resizebox{\textwidth}{!}{
\begin{tabular}{ll c S[table-format=2.1, table-align-text-post=false]
                     S[table-format=2.1, table-align-text-post=false]
                     S[table-format=2.1, table-align-text-post=false]}
\toprule[1.5pt]
\textbf{Model} & \textbf{Approach} & \textbf{PCGrad}  & {\textbf{NER (5 DS)}} & {\textbf{TC (16 DS)}} & {\textbf{Average}} \\
\midrule[1.pt]
\multirow{6}{*}{\small MiniLM} & \texttt{Individual} & $-$ &\textcolor{gray}{77.1} & \textcolor{gray}{63.9} & \textcolor{gray}{70.5} \\
\hhline{~-----}
 & \texttt{Base model} & $-$ & 76.4 & 55.1 & 65.8\\
& \texttt{MTPF} & \textcolor{red}{\ding{55}} & 76.4 & 63.6 \small\textcolor{green}{+8.5} & 70.0 \small\textcolor{green}{+4.2}\\
& \texttt{MTPF} & \textcolor{green}{\ding{51}} & 76.5 \small\textcolor{green}{+0.1} & 63.7 \small\textcolor{green}{+8.6} & 70.1 \small\textcolor{green}{+4.3}\\
& \ourmethod{}  & \textcolor{red}{\ding{55}} & 77.1 \small\textcolor{green}{+0.8} & 63.9 \small\textcolor{green}{+8.8} & 70.5 \small\textcolor{green}{+4.7}\\
& \ourmethod{}  & \textcolor{green}{\ding{51}} & 77.1 \small\textcolor{green}{+0.8} & 64.1 \small\textcolor{green}{+9.0} & 70.6 \small\textcolor{green}{+4.9}\\
\midrule[0.8pt]
\multirow{6}{*}{\small DistilBERT} & \texttt{Individual} & $-$ &\textcolor{gray}{77.7} & \textcolor{gray}{64.4} & \textcolor{gray}{71.1} \\
\hhline{~-----}
 & \texttt{Base model} & $-$ & 76.9& 60.5 & 68.7\\
& \texttt{MTPF} & \textcolor{red}{\ding{55}} & 77.2 \small\textcolor{green}{+0.3} & 64.0 \small\textcolor{green}{+3.5} & 70.6 \small\textcolor{green}{+1.9}\\
& \texttt{MTPF} & \textcolor{green}{\ding{51}} & 77.2 \small\textcolor{green}{+0.3} & 63.9 \small\textcolor{green}{+3.4} & 70.6 \small\textcolor{green}{+1.9}\\
& \ourmethod{} & \textcolor{red}{\ding{55}} & 77.6 \small\textcolor{green}{+0.7} & 64.4 \small\textcolor{green}{+3.9} & 71.0 \small\textcolor{green}{+2.3}\\
& \ourmethod{} & \textcolor{green}{\ding{51}} &77.7\small\textcolor{green}{+0.8} & 64.4 \small\textcolor{green}{+3.9} & 71.1 \small\textcolor{green}{+2.4}\\
\bottomrule[1.5pt]
\end{tabular}
}
\end{minipage}
\hfill
\begin{minipage}[t!]{0.3\textwidth}
\centering
\caption{\textbf{Comparison of downstream performance across different pre-finetuning strategies and the base model}. NER results represent an average over 5 datasets (DS), while text classification (TC) results represent an average over 16 datasets. Result of each dataset is given in \cref{tab:minilm_ner,tab:minilm_tc}. Results for individual pre-finetuning are grayed out, as the resulting two models are not compatible with our single backbone scheme. \textcolor{green}{Gain} and \textcolor{red}{loss} are compared with the base model.}
\label{tab:main_res}
\end{minipage}
\end{table*}

\begin{figure*}[t!]
    \centering
    \begin{subfigure}[b]{0.23\textwidth}
        \centering
        \includegraphics[width=\textwidth]{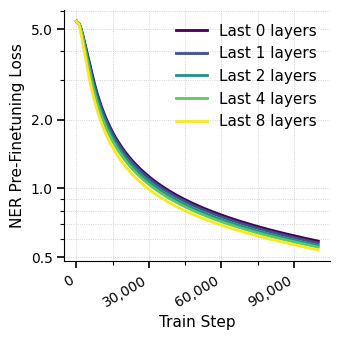}
        \caption{NER Loss vs. Train Step}
        \label{fig:ner_loss_vs_train_step}
    \end{subfigure}
    \hfill
    \begin{subfigure}[b]{0.23\textwidth}
        \centering
        \includegraphics[width=\textwidth]{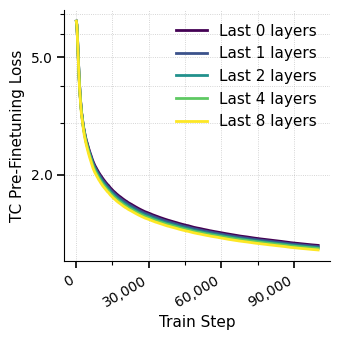}
        \caption{TC Loss vs. Train Step}
        \label{fig:tc_loss_vs_train-step}
    \end{subfigure}
    \hfill
    \begin{subfigure}[b]{0.23\textwidth}
        \centering
        \includegraphics[width=\textwidth]{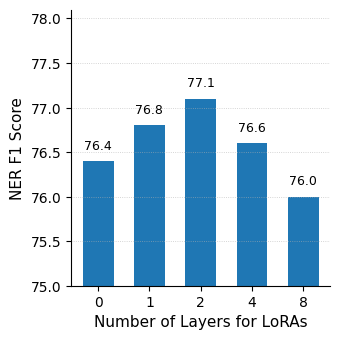}
        \caption{NER Downstream Perf.}
        \label{fig:comparison_ner_performance}
    \end{subfigure}
    \hfill
    \begin{subfigure}[b]{0.23\textwidth}
        \centering
        \includegraphics[width=\textwidth]{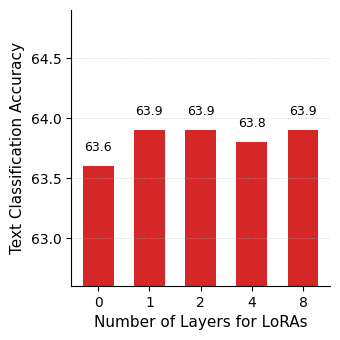}
        \caption{TC Downstream Perf.}
        \label{fig:comparison_tc_performance}
    \end{subfigure}
    \caption{\textbf{Comparison of applying task-primary LoRAs (TPL) to varying numbers of final layers.} (a) and (b) show pre-finetuning loss over training steps for different numbers of layers augmented with TPL. (c) and (d) present downstream performance across both task families under varying numbers of TPL-applied layers.}
    \label{fig:layers_analysis}
\end{figure*}

\section{\underline{M}ulti-\underline{T}ask \underline{P}re-\underline{F}inetuning with \underline{T}ask-\underline{P}rimary \underline{L}oRAs (\ourmethod)}
\label{sec:mtl}
\paragraph{Multi-Task Pre-Finetuning.}
As discussed in \cref{sec:insight}, pre-finetuning strategies for NER and text classification impose conflicting requirements on token embeddings, leading to incompatible optimization directions. However, our deployment setting requires a single shared encoder (see \cref{fig:setting}), making it necessary to merge these strategies through multi-task pre-finetuning.

Unlike traditional multi-task learning~\citep{liu-etal-2015-representation,ruder2017overview,liu-etal-2019-multi}, which typically involves multiple datasets across domains to improve data diversity, our individual pre-finetuning approach already relies on large-scale, general-purpose datasets. Instead of seeking complementary benefits, our goal is to resolve the incompatibility between the pre-finetuning strategies for NER and text classification to enable unified encoder optimization.

\paragraph{Task-Primary LoRAs.}
Since NER and text classification favor different token embedding characteristics, we extend the encoder with two groups of LoRA modules. Unlike the typical use of LoRA—where the backbone is frozen and only the LoRA parameters are updated—we allow joint optimization: each LoRA is updated solely with the loss of its associated task, while the backbone encoder is optimized by both loss functions, serving as shared parameters (see \cref{fig:arch}).

After multi-task pre-finetuning, the backbone encoder is deployed as the central model, and the task-primary LoRAs are distributed to applications for downstream adaptation. These LoRAs can be used either as initialization for downstream LoRA adaptation or directly for inference with linear probing. We refer to these pre-finetuning LoRA parameters as \textit{task-primary LoRAs} (TPL).

Task-primary LoRAs are inspired by multi-task learning approaches such as \texttt{Cross-Stitch}~\citep{7780802}, which trains task-specific networks alongside a shared representation network. However, in our case, attaching task-specific networks or duplicating encoder layers does not meet the system requirement of maintaining a single backbone with modular adapters. Additionally, \texttt{Bert-and-PALs}\citep{pmlr-v97-stickland19a} introduces task-specific adapters into all attention layers for multi-task learning, but their approach directly targets downstream tasks and focuses on minimizing parameter count. In contrast, \ourmethod{} focuses on pre-finetuning to improve adaptability across downstream tasks. As we will discuss later, applying task-primary LoRAs only to the last few layers is key to achieving this adaptability.

\subsection{Results}
Next, we demonstrate the effectiveness of \ourmethod{} on two parameter-efficient models: MiniLM~\citep{NEURIPS2020_3f5ee243} and DistilBERT~\citep{sanh2019distilbert}. We compare \ourmethod{} against the pre-trained base model, individually pre-finetuned models, and multi-task pre-finetuning (\texttt{MTPF}) without task-primary LoRAs. Additionally, we include \texttt{PCGrad}~\citep{NEURIPS2020_3fe78a8a}, a gradient surgery method commonly used in multi-task learning to mitigate gradient conflicts by projecting gradients into orthogonal spaces. The results are presented in \cref{tab:main_res}. We observe the following:
\textbf{1)} \texttt{MTPF} without task-primary LoRAs benefits from pre-finetuning and outperforms the base model on downstream tasks. However, due to task interference, its improvements are smaller than those achieved by individually pre-finetuned models.
\textbf{2)} When equipped with task-primary LoRAs, \ourmethod{} successfully combines the strengths of both pre-finetuning strategies, achieving performance comparable to individual pre-finetuning while maintaining a single-backbone model suitable for deployment. Additionally, as discussed in \cref{sec:pre-finetuneing}, pre-finetuning for NER yields substantial gains in low-resource settings (e.g., \textcolor{green}{$+8.4\%$} F1 when using only $10\%$ of the original training data). We observe that \texttt{MTPF-TPL} preserves this improvement (see \cref{app:limited_data}).
\textbf{3)} While \texttt{PCGrad} slightly improves performance by reducing gradient conflicts, it does not mitigate task interference as effectively as task-primary LoRAs.

In \cref{tab:main_res}, task-primary LoRAs are applied only to the last two transformer layers, which we find optimal for both MiniLM and DistilBERT in our deployment setting. Constraining task-primary LoRAs to the final layers appears crucial for adaptability to downstream tasks, which we discuss below.

\begin{table}[t]
\centering
\resizebox{\columnwidth}{!}{
\begin{tabular}{l S[table-format=2.1, table-align-text-post=false]
                   S[table-format=2.1, table-align-text-post=false]
                   S[table-format=2.1, table-align-text-post=false]}
\toprule[1.5pt]
\textbf{Model} & {\textbf{NER (5 DS)}} & {\textbf{TC (16 DS)}} & {\textbf{Average}} \\
\midrule[1.pt]
\texttt{Individual} & 77.1 & 63.9 & 70.5\\
\midrule
Base model & 76.4 & 55.1 & 65.8\\
PF-L (two layers) & 76.3 \small\textcolor{red}{-0.1} & 60.2 \small\textcolor{green}{+5.1} & 68.3 \small\textcolor{green}{+2.5} \\
PF-L (all layers) & 76.6 \small\textcolor{green}{+0.2} & 62.4 \small\textcolor{green}{+7.3} & 69.5 \small\textcolor{green}{+3.7}\\
MTPF-TPL & 77.1 \small\textcolor{green}{+0.7} & 63.9 \small\textcolor{green}{+8.8} & 70.5 \small\textcolor{green}{+4.7}\\
\bottomrule[1.5pt]
\end{tabular}
}
\caption{\textbf{Comparison of downstream performance between classical LoRA fine-tuning (denoted as PF-L) and ours MTPF-TPL.} NER results report the average F1 score over 5 datasets (DS), while text classification (TC) results report the average accuracy over 16 datasets. \textcolor{green}{Gain} and \textcolor{red}{Loss} are measured relative to the base model. The model architecture used is MiniLM.}
\label{tab:lora_only}
\end{table}
\paragraph{Applying Task-Primary LoRAs to the Last Few Transformer Layers Improves Adaptability.}
As shown in \cref{fig:layers_analysis}, pre-finetuning loss consistently decreases as more transformer layers of MiniLM are equipped with task-primary LoRAs. However, downstream performance on NER peaks when LoRAs are applied only to the last two layers. This phenomenon is unlikely to result from overfitting, given the large scale of our pre-finetuning dataset (\cref{sec:pre-finetuneing}) and the strong downstream results of individually pre-finetuned models (\cref{tab:main_res}).

In our adaptation setup (\cref{sec:adapt}), task-primary LoRAs serve as a partial initialization for LoRA modules (applied to all layers) that are further fine-tuned on NER sub-tasks. In contrast, for text classification, task-primary LoRAs remain fixed, and only the linear classifier is updated. Since the performance degradation does not appear for text classification, we hypothesize that once the pre-finetuning conflict is mitigated, initializing more transformer layers with random LoRA parameters—rather than task-primary LoRAs—may improve LoRA adaptation on downstream sub-tasks.

\paragraph{Freezing the backbone and only fine-tune LoRAs.} We also compare our method with classical LoRA fine-tuning (denoted as PF-L) for different task families, following the pipeline that LoRAs are first pre-finetuned and then further adapted for downstream tasks. While this strategy avoids interference between task families, its performance remains below that of individual full-parameter pre-finetuning as shown in \cref{tab:lora_only}, likely due to the limited number of trainable parameters. Applying PF-L to all layers performs better than restricting them to the last two (our \texttt{MTPF-TPL}'s setup). While further increasing LoRA rank might close the gap to individual pre-finetuning, this would also scale parameter cost linearly with the number of applications—potentially dozens or hundreds on today’s smartphones. By contrast, \texttt{MTPF-TPL}’s shared backbone approach achieves better parameter efficiency while maintaining strong performance, making it more suitable for on-device deployment.



\section{Conclusion}
\label{sec:conclusion}
In this work, we present a deployment setup for NLP tasks on mobile platforms and explore pre-finetuning strategies for two key task families: named entity recognition (NER) and text classification. We demonstrate that pre-finetuning improves downstream performance for both task families but also identify interference when the strategies are combined. To address this, we propose a multi-task pre-finetuning framework with task-primary LoRAs that effectively integrates both approaches, resulting in a model and modular adapters compatible with our deployment requirements.

\section*{Limitations}
Our study focuses on demonstrating the effectiveness of multi-task pre-finetuning with task-primary LoRAs in a controlled setting. While the results show consistent improvements across diverse NER and text classification tasks, there remain several avenues for future exploration.

First, we conduct experiments exclusively using English pre-trained models and English-language datasets. This is a common first step in NLP system development, but extending our approach to multilingual settings remains an important direction for future work.

Second, for consistency, we adopt the same task-primary LoRA configuration (e.g., rank and number of augmented layers) across both task families. While our findings already yield strong results under this unified design, task-specific adapter configurations could potentially offer further improvements in downstream performance or efficiency.


\bibliography{custom}

\appendix
\section{Details of Adaptation}
\label{app:adapt}
For NER sub-tasks, we apply LoRA to the key and query matrices of all attention layers as well as to the MLP layers, using a rank of 32 and an $\alpha$ of 64. Since the linear classifier is randomly initialized, we first freeze the encoder and train the classifier alone for 10 epochs to warm it up. We then jointly fine-tune both the encoder and classifier for an additional 30 epochs. This two-stage procedure consistently yields better performance than training without warm-up in our experiments. We set the batch size to 64 and the learning rate to $2 \times 10^{-5}$, using the AdamW optimizer~\citep{loshchilov2018decoupled} with a weight decay of 0.01.

For text classification sub-tasks, we follow the standard procedure of \texttt{MTEB}, performing logistic regression on top of the text embeddings extracted by the encoder.
\section{Details of Pre-Finetuning}
\label{app:pf}
For text classification sub-tasks, we adopt weakly supervised contrastive learning with text pairs as the pre-finetuning strategy. In this setting, both the semantic similarity between anchor texts and their positive pairs, as well as the diversity of all text samples, are crucial for optimization. We combine multiple datasets pre-processed by \texttt{SentenceTransformer}\footnote{\url{https://sbert.net/docs/sentence_transformer/dataset_overview.html\#datasets-on-the-hugging-face-hub}} to maximize semantic coverage and similarity. The datasets used include:
\texttt{all-nli}, \texttt{quora-duplicates}, \texttt{stackexchange-duplicates}, \texttt{wikihow}, \texttt{xsum}, \texttt{s2orc}, \texttt{wikianswers-duplicates}, \texttt{agnews}, \texttt{npr}, \texttt{specter}, \texttt{simple-wiki}, \texttt{altlex}, \texttt{ccnews}, \texttt{sentence-compression}, \texttt{flickr30k-captions}, and \texttt{amazon-reviews}.

We apply an in-batch negatives strategy with a batch size of 1024 to ensure sufficient diversity of negative pairs. Additionally, we employ a hierarchical data sampling strategy, where we first sample a dataset and then sample a batch from that dataset. This approach consistently outperforms global random sampling~\cite{gunther-etal-2023-jina,gunther2023jina}. We use the AdamW optimizer with a learning rate of $2 \times 10^{-5}$ and train for 100K iterations.

For NER pre-finetuning, we generate pseudo labels by applying $k$-means clustering to token embeddings extracted from the \texttt{NuNER} model. To improve the clustering quality, we compute the average of sub-token embeddings over a word and run clustering on the resulting word embedding. Since \texttt{NuNER} was originally trained to recognize 200K concepts~\citep{bogdanov-etal-2024-nuner} with overlaps of the concepts, we conduct a grid search over $\{50, 100, 200, 500, 1000\}$ clusters and find that 200 yields the best downstream performance. We similarly grid search the batch size over $\{16, 32, 64, 128, 256, 512\}$ and select 256 based on downstream results. We adopt the same learning rate and training iterations as used for text classification pre-finetuning.

\section{Detailed Downstream Performance on Sub-Tasks}
\Cref{tab:pft_ner,tab:pft_tc} provide detailed breakdowns of the downstream performance reported in \cref{tab:pft}. Similarly, \cref{tab:minilm_ner,tab:minilm_tc,tab:distilbert_ner,tab:distilbert_tc} present the breakdowns corresponding to the results in \cref{tab:main_res}.

\begin{figure}[t]
\centering
\begin{tcolorbox}[colback=gray!5, colframe=gray!50, width=\linewidth, boxrule=0.5pt, arc=2pt]
\small
\textbf{Original Sentence:} \\
\emph{During his visit to \ent{Slovenia}, \ent{Kwasniewski} is also scheduled to meet Prime Minister.} \\[0.8em]
\textbf{Perturbed Sentences:} \\
\emph{During his visit to \ent{Iran}, \ent{Kim Yoon-man} is also...}\\[0.1em]
\emph{During his visit to \ent{Central African Republic}, \ent{Marc Cohen} is also...}\\[0.1em]
\emph{During his visit to \ent{Brno}, \ent{Blaise Compaore} is also...} \\[0.1em]
\emph{During his visit to \ent{UK}, \ent{Eisuke Sakakibara} is also...}
\end{tcolorbox}
\caption{\textbf{Example of entity replacement}. Location and person entities (highlighted) have been substituted with random entities of the same type.}
\label{fig:entity-perturbation}
\end{figure}
\begin{table}[t]
\centering
\resizebox{\columnwidth}{!}{
\begin{tabular}{l S[table-format=2.1, table-align-text-post=false]
                   S[table-format=2.1, table-align-text-post=false]
                   S[table-format=2.1, table-align-text-post=false]}
\toprule[1.5pt]
\textbf{Model} & {\textbf{NER (5 DS)}} & {\textbf{TC (16 DS)}} & {\textbf{Average}} \\
\midrule[1.pt]
MiniLM & 76.4 & 55.1 & 65.8\\
all-MiniLM & 74.5 \small\textcolor{red}{-1.9} & 68.2 \small\textcolor{green}{+13.1} & 71.4 \small\textcolor{green}{+5.6} \\
\midrule
RoBERTa-base & 79.6 & 58.7 & 69.1\\
NUNER & 80.3 \small\textcolor{green}{+0.7} & 58.1 \small\textcolor{red}{-0.6} & 69.2 \small\textcolor{green}{+0.1} \\
\bottomrule[1.5pt]
\end{tabular}
}
\caption{\textbf{Comparison of downstream performance across different open-source pre-finetuned models and their base models.} NER results report the average F1 score over 5 datasets (DS), while text classification (TC) results report the average accuracy over 16 datasets. \textcolor{green}{Gain} and \textcolor{red}{Loss} are measured relative to the base model.}
\label{tab:open_source_pft}
\end{table}

\begin{table*}[t!]
\centering
\resizebox{0.75\textwidth}{!}{ 
\begin{tabular}{lcccccc}
\toprule[1.5pt]
\textbf{Model}  & \textbf{CoNLL} & \textbf{OntoNotes5} & \textbf{MIT Restaurant} &\textbf{BioNLP} &\textbf{CrossNER} & \textbf{Average} \\
\midrule[.8pt]
Base model & 88.7 & 84.6 & 76.7 & 68.1 & 64.1 & 76.4\\
PF for NER & \bftab 89.4 & \bftab 84.9 & \bftab 77.1 & \bftab 68.9 & \bftab 65.2 & \bftab 77.1\\
PF for TC &  87.3& 83.5& 75.4 & 67.1 & 60.3 & 74.6\\
\bottomrule[1.5pt]
\end{tabular}
}
\caption{\textbf{Comparison of downstream performance of different pre-finetuning (PF) approaches and the base model across NER sub-tasks.} Evaluation metric is F1. TC stands for text classification. Model architecture is MiniLM. The best results are bold.}
\label{tab:pft_ner}
\end{table*}

\begin{table*}[t!]
\centering
\resizebox{\textwidth}{!}{ 
\begin{tabular}{lccccccccccccccccc}
\toprule[1.5pt]
\textbf{Model}  & \textbf{Bank.} & \textbf{Emo.} & \textbf{ACF} &\textbf{MI} &\textbf{TS} & \textbf{TC} & \textbf{News} & \textbf{Patent} & \textbf{FPB} & \textbf{FE} & \textbf{Imdb} & \textbf{Arxiv} & \textbf{DB.} & \textbf{TTS.} & \textbf{YRF} & \textbf{TCon.} & \textbf{Avg.} \\
\midrule[.8pt]
\small Base model & 58.8 & 29.8 & \bftab 72.9 & 54.4  & \bftab 46.3 & \bftab 69.7& \bftab 74.1 & 25.7 & 61.4 & 57.6 & 58.6 & 33.7 & 80.5 & 47.0 & 46.4 & 65.0 & 55.1\\
\small PF for NER & 53.4 & 27.9& 66.2& 51.8 &44.3 &66.6 &71.9 & 27.1 & 54.6 & 56.3 & 57.3 & 39.5 & 83.5 & 52.5 & 43.6 & 65.5 & 53.9\\
\small PF for TC & \bftab 79.5& \bftab 40.2 & 69.9 & \bftab 64.2  & 43.5 & 63.0 & 73.5 & \bftab 36.6 & \bftab 72.7 & \bftab 59.6 & \bftab 85.7 & \bftab 63.0 & \bftab 84.5 & \bftab 67.4 & \bftab 53.1 & \bftab 66.1 & \bftab 63.9 \\
\bottomrule[1.5pt]
\end{tabular}
}
\caption{\textbf{Comparison of downstream performance of different pre-finetuning (PFT) approaches and the base model on text classification (TC) sub-tasks.} Accuracy is used as the evaluation metric. The model architecture is MiniLM. Best results are highlighted in bold. Dataset names are abbreviated as follows: Bank.: Banking77, Emo.: Emotion, ACF: AmazonCounterfactual, MI: MassiveIntent, TS: TweetSentiment, TC: ToxicChat, FPB: FinancialPhrasebank, FE: FrenkEn, DB: DBpedia, TTS: TweetTopicSingle, YRF: YelpReviewFull, TCon.: ToxicConversation.}
\label{tab:pft_tc}
\end{table*}

\begin{table*}[t!]
\centering
\resizebox{0.8\textwidth}{!}{ 
\begin{tabular}{lccccccc}
\toprule[1.5pt]
\textbf{Approach}  & \textbf{PCGrad} & \textbf{CoNLL} & \textbf{OntoNotes5} & \textbf{MIT Restaurant} &\textbf{BioNLP} &\textbf{CrossNER} & \textbf{Average} \\
\midrule[.8pt]
\texttt{MTPF} &\textcolor{red}{\ding{55}} & 88.5 & 84.2 & 76.8 & 67.1 & 65.6 & 76.4\\
\texttt{MTPF} &\textcolor{green}{\ding{51}} & 88.7 & 84.3 & 77.2 & 67.0 & 65.3 & 76.5\\
\ourmethod{} & \textcolor{red}{\ding{55}} &89.1& \bftab 84.6& \bftab 77.4 & 68.1 &\bftab  66.3 & \bftab 77.1\\
\ourmethod{} & \textcolor{green}{\ding{51}} &\bftab 89.2& 84.5& \bftab 77.4 & \bftab 68.2 & 66.1 & \bftab 77.1\\
\bottomrule[1.5pt]
\end{tabular}
}
\caption{\textbf{Comparison of downstream performance of multi-task pre-finetuning (\texttt{MTPF}) with or without task-primary LoRAs (\texttt{TPL}) across NER sub-tasks.} Model architecture is \textbf{\textit{MiniLM}}. Evaluation metric is F1. The best results are bold.}
\label{tab:minilm_ner}
\end{table*}

\begin{table*}[t!]
\centering
\resizebox{\textwidth}{!}{ 
\begin{tabular}{lcccccccccccccccccc}
\toprule[1.5pt]
\textbf{Approach}  &\textbf{PCGrad} & \textbf{Bank.} & \textbf{Emo.} & \textbf{ACF} &\textbf{MI} &\textbf{TS} & \textbf{TC} & \textbf{News} & \textbf{Patent} & \textbf{FPB} & \textbf{FE} & \textbf{Imdb} & \textbf{Arxiv} & \textbf{DB.} & \textbf{TTS.} & \textbf{YRF} & \textbf{TCon.} & \textbf{Avg.} \\
\midrule[.8pt]
\small \texttt{MTPF} & \textcolor{red}{\ding{55}}& 78.3& 40.0 & \bftab 70.3 & 63.6 & 45.1 & 61.1 & \bftab 74.1 &\bftab 36.7 & 69.8 & 58.0 & 84.3 & \bftab 63.3 & 86.1 & \bftab 67.9 & 52.5 & 67.2 & 63.6\\
\small \texttt{MTPF} & \textcolor{green}{\ding{51}}& 78.5 & 39.6 & 69.5 & 64.0 & \bftab 45.5 & 62.3 & 73.3 & 36.4 & 71.7 & 58.0 & \bftab 85.7 & 63.0 & \bftab 86.2 & 66.7 & 53.0 & 66.7 & 63.7 \\
\small \ourmethod{} &\textcolor{red}{\ding{55}}& \bftab 79.5 & \bftab 40.2 & 69.9 & 64.2 & 43.5 &\bftab 63.0 & 73.5 &36.6 &72.7 &\bftab 59.6 & \bftab 85.7 & 63.0 & 84.5 & 67.4 & 53.1 & 66.1 & 63.9\\
\small \ourmethod{} &\textcolor{green}{\ding{51}}& 79.4 & \bftab 40.2 & 70.1 & \bftab 64.3 & 45.0 & 62.1 & 73.9 & 36.5 & \bftab 73.0 & 59.3 & 85.5 & 62.6 & 84.9 & 67.2 &\bftab 53.4 & \bftab 67.6 & \bftab 64.1 \\
\bottomrule[1.5pt]
\end{tabular}
}
\caption{\textbf{Comparison of downstream performance of multi-task pre-finetuning (\texttt{MTPF}) with and without task-primary LoRAs (TPL) on text classification (TC) sub-tasks.} Accuracy is used as the evaluation metric. The model architecture is \textbf{\textit{MiniLM}}. Best results are highlighted in bold. Dataset names are abbreviated as follows: Bank.: Banking77, Emo.: Emotion, ACF: AmazonCounterfactual, MI: MassiveIntent, TS: TweetSentiment, TC: ToxicChat, FPB: FinancialPhrasebank, FE: FrenkEn, DB: DBpedia, TTS: TweetTopicSingle, YRF: YelpReviewFull, TCon.: ToxicConversation.}
\label{tab:minilm_tc}
\end{table*}

\begin{table*}[t!]
\centering
\resizebox{0.8\textwidth}{!}{ 
\begin{tabular}{lccccccc}
\toprule[1.5pt]
\textbf{Approach}  & \textbf{PCGrad} & \textbf{CoNLL} & \textbf{OntoNotes5} & \textbf{MIT Restaurant} &\textbf{BioNLP} &\textbf{CrossNER} & \textbf{Average} \\
\midrule[.8pt]
\texttt{MTPF} &\textcolor{red}{\ding{55}} & 88.0 & 84.1 & 76.4 & 68.7 & 68.7 & 77.2\\
\texttt{MTPF} &\textcolor{green}{\ding{51}} & 88.1 &\bftab 84.2 & 76.6 & 68.8 & 68.3 & 77.2\\
\ourmethod{} & \textcolor{red}{\ding{55}} & 88.6 &\bftab 84.2 & 77.0 & 69.0 & 68.9 & 77.6\\
\ourmethod{} & \textcolor{green}{\ding{51}} & \bftab 88.8 & \bftab 84.2 & \bftab 77.3 & \bftab 69.2 & \bftab 69.0& \bftab 77.7\\
\bottomrule[1.5pt]
\end{tabular}
}
\caption{\textbf{Comparison of downstream performance of multi-task pre-finetuning (\texttt{MTPF}) with or without task-primary LoRAs (\texttt{TPL}) across NER sub-tasks.} Model architecture is \textbf{\textit{DistilBERT}}. Evaluation metric is F1. The best results are bold.}
\label{tab:distilbert_ner}
\end{table*}

\begin{table*}[t!]
\centering
\resizebox{\textwidth}{!}{ 
\begin{tabular}{lcccccccccccccccccc}
\toprule[1.5pt]
\textbf{PFT}  &\textbf{PCGrad} & \textbf{Bank.} & \textbf{Emo.} & \textbf{ACF} &\textbf{MI} &\textbf{TS} & \textbf{TC} & \textbf{News} & \textbf{Patent} & \textbf{FPB} & \textbf{FE} & \textbf{Imdb} & \textbf{Arxiv} & \textbf{DB.} & \textbf{TTS.} & \textbf{YRF} & \textbf{TCon.} & \textbf{Avg.} \\
\midrule[.8pt]
\small \texttt{MTPF} & \textcolor{red}{\ding{55}}& 78.6 & 42.8 & 69.6 & 64.6 & 45.4& 62.7& 71.5 & 35.9 & 71.3 & 58.1 & 85.7 & 64.7 & \bftab 87.9 & \bftab 68.0 & 51.4 & 66.2 & 64.0\\
\small \texttt{MTPF} & \textcolor{green}{\ding{51}}& 78.5 & 43.5 & 70.3& 64.4 & 44.9 & \bftab 63.0 & 71.2& 35.5 & 71.5 & 58.0 & 84.9 & 64.4 & 87.8 & 67.5 &51.4 & 65.8 & 63.9\\
\small \ourmethod{} &\textcolor{red}{\ding{55}}&\bftab 79.4 & 43.6 & 70.1 & 65.5 & 46.3 & 62.2 & 67.5 & \bftab 36.3 & 75.4 & \bftab 58.4 & 85.9& \bftab 65.5 & 86.4 &\bftab 68.0 & \bftab 53.0 & 66.7 & \bftab 64.4\\
\small \ourmethod{} &\textcolor{green}{\ding{51}}& \bftab 79.4 & \bftab 44.4 & 70.3 & \bftab 65.6 & \bftab 46.8 & 61.2 & 67.7 & 36.2 & \bftab 75.6 & 58.3 &\bftab 86.0 & 65.4 & 86.6 & 67.7 & 52.9 & \bftab 66.8 & \bftab 64.4\\
\bottomrule[1.5pt]
\end{tabular}
}
\caption{\textbf{Comparison of downstream performance of multi-task pre-finetuning (\texttt{MTPF}) with and without task-primary LoRAs (\texttt{TPL}) on text classification (TC) sub-tasks.} Accuracy is used as the evaluation metric. The model architecture is \textbf{\textit{DistilBERT}}. Best results are highlighted in bold. Dataset names are abbreviated as follows: Bank.: Banking77, Emo.: Emotion, ACF: AmazonCounterfactual, MI: MassiveIntent, TS: TweetSentiment, TC: ToxicChat, FPB: FinancialPhrasebank, FE: FrenkEn, DB: DBpedia, TTS: TweetTopicSingle, YRF: YelpReviewFull, TCon.: ToxicConversation.}
\label{tab:distilbert_tc}
\end{table*}
\section{Downstream Performance of NER Sub-Tasks with Limited Data}
\label{app:limited_data}

NER tasks typically require costly human annotation by domain experts, making it challenging to obtain large-scale labeled datasets in many real-world scenarios. To assess the benefit of pre-finetuning for NER under low-resource conditions, we evaluate downstream adaptation when only a small fraction of the original labeled data is available.

As shown in \cref{tab:limited_data}, pre-finetuning significantly enhances downstream performance, and this improvement becomes more pronounced as the amount of labeled data decreases. When only $10\%$ of the original training data is used, pre-finetuned MiniLM achieves an average F1 gain of \textcolor{green}{$+8.4\%$} over its base model, while pre-finetuned DistilBERT achieves \textcolor{green}{$+5.3\%$}. These results highlight the value of pre-finetuning in low-resource NER scenarios, where model generalization from limited supervision is critical.

Importantly, our proposed \texttt{MTPF-TPL} framework preserves this advantage while providing a single unified pre-finetuned model that supports both NER and text classification. This enables efficient deployment without sacrificing the gains of pre-finetuning, even in data-scarce settings.

\begin{table*}[t!]
\centering
\resizebox{0.98\textwidth}{!}{ 
\begin{tabular}{lclcccccc}
\toprule[1.5pt]
\textbf{Architecture }& \textbf{Data Portion}& \textbf{Approach}  & \textbf{CoNLL} & \textbf{OntoNotes5} & \textbf{MIT Restaurant} &\textbf{BioNLP} &\textbf{CrossNER} & \textbf{Average} \\
\midrule[.8pt]
\multirow{12}[4]{*}{MiniLM}&\multirow{4}{*}{$10\%$} &Base model & 66.5& 71.1 & 29.4 & 45.6 & 23.7 & 47.3\\
& &PF for NER & 67.3 & 73.8 & 45.2 & 51.4 & 35.6 & 55.7\\
& &\texttt{MTPF} & 65.6 & 73.3 & 35.3 & 51.4& 35.3 & 52.2\\
& &\texttt{MTPF-TPL} &\bftab 67.5 & \bftab 74.9 & \bftab 47.2 & \bftab 51.9 & \bftab 37.9 & \bftab 55.9\\
\cmidrule(l){2-9}
&\multirow{4}{*}{$20\%$} &Base model & 83.3 & 77.9 & 55.0 & 53.0 & 37.7 & 62.4\\
&&PF for NER & 86.2 & 79.3 & 58.3 & \bftab 59.7 & 47.8 & 66.3\\
&&\texttt{MTPF} & 86.1 & 79.1 & 56.4 & 59.0 & 45.8 & 65.3\\
&&\texttt{MTPF-TPL} & \bftab 86.7 &\bftab 80.3 &\bftab 60.8 & 59.6 &\bftab 48.7 &\bftab 67.2\\
\cmidrule(l){2-9}
&\multirow{4}{*}{$50\%$} &Base model & 87.3 & 82.6 & 71.6 & 64.7 & 53.9 & 72.0\\
&&PF for NER &\bftab 89.1 & \bftab 83.3 & 73.1 & \bftab 66.3 & 58.5 & 74.1\\
&&\texttt{MTPF} & 87.8 & 82.7 & 71.8 & 65.0 & 59.4 & 73.4\\
&&\texttt{MTPF-TPL} & 88.3 & \bftab 83.3 & \bftab 75.0 & 65.4 & \bftab 61.2 & \bftab 74.6\\
\midrule[.8pt]
\multirow{12}[4]{*}{DistilBERT}&\multirow{4}{*}{$10\%$} &Base model & 65.7 & 72.9 & 27.4 & 50.1 & 38.0 & 50.8\\
& &PF for NER & 67.5 & 74.0 & \bftab 42.8 & 52.2 & \bftab 43.7 & \bftab 56.1\\
& &\texttt{MTPF} & 64.4 & 72.0 & 39.1 & 51.1 & 41.7 & 53.7\\
& &\texttt{MTPF-TPL} & \bftab 69.5 &\bftab 74.2 &39.9 &\bftab 52.6 & 42.4 & 55.7  \\
\cmidrule(l){2-9}
&\multirow{4}{*}{$20\%$} &Base model &80.9 & 78.6 & 55.4 & 60.2 & 50.3 & 65.6\\
&&PF for NER &\bftab 85.0 & \bftab 79.7& 53.2 &\bftab 61.5 & 54.3 & \bftab 66.8\\
&&\texttt{MTPF} & 80.9 & 79.4 & \bftab 56.9 & 60.7 & 53.4 & 66.3\\
&&\texttt{MTPF-TPL} & 82.6 & 79.6 & 55.6 & \bftab 61.5 & 54.7 &\bftab 66.8\\
\cmidrule(l){2-9}
&\multirow{4}{*}{$50\%$} &Base model & 87.0 & 82.6 & 70.6 & \bftab 67.1 & 63.3 & 74.1\\
&&PF for NER & \bftab 87.8 & \bftab 83.3 & 72.1 & 67.0 & 64.6 & \bftab 75.0\\
&&\texttt{MTPF} & 86.8 & 82.5 & 72.8 & 66.4 & 63.5 & 74.4\\
&&\texttt{MTPF-TPL} &87.3 & 82.6 & \bftab 73.0 & 67.0 & \bftab 64.7 & 74.8\\
\bottomrule[1.5pt]
\end{tabular}
}
\caption{\textbf{Comparison of downstream performance of different approaches under data scarcity across NER sub-tasks.} The evaluation metric is F1. The data portion indicates the percentage of the original training data used for fine-tuning. The best results are shown in bold.}
\label{tab:limited_data}
\end{table*}
\section{Reproducing the Interference Between Pre-Finetuning Strategies Using Open-Source Models}
\label{app:open_source}

In \cref{sec:insight}, we demonstrate the interference between our implemented pre-finetuning strategies. Here, we show that this is a general phenomenon observable with open-source pre-finetuned models. Specifically, we compare \texttt{NuNER}~\citep{bogdanov-etal-2024-nuner} with its base model \texttt{RoBERTa-base}~\citep{liu2019roberta}, and \texttt{all-MiniLM}\footnote{\url{https://huggingface.co/sentence-transformers/all-MiniLM-L12-v2}} with its base model \texttt{MiniLM}~\citep{NEURIPS2020_3f5ee243}.

As shown in \cref{tab:open_source_pft}, \texttt{NuNER}, pre-finetuned for NER, exhibits reduced adaptability to text classification sub-tasks compared to \texttt{RoBERTa-base}. Conversely, \texttt{all-MiniLM}, pre-finetuned for text embedding tasks, shows worse adaptability to NER compared to its base model \texttt{MiniLM}.

\end{document}